%% file: paper.tex
\documentclass[letterpaper, 10 pt, conference]{ieeeconf}

\RequirePackage[normalem]{ulem} 
\RequirePackage{color}\definecolor{RED}{rgb}{1,0,0}\definecolor{BLUE}{rgb}{0,0,1} 

\usepackage[pdftex]{graphicx}
\graphicspath{{images}}
\DeclareGraphicsExtensions{.pdf,.jpeg,.png}

\usepackage{amsmath}
\usepackage{array}

\usepackage[font=small]{caption}
\usepackage{subcaption}
\usepackage{lipsum}
\usepackage{setspace}
\let\labelindent\relax
\usepackage{enumitem}
\usepackage{url}
\usepackage{amssymb}
\usepackage{bm}
\usepackage{graphicx}
\usepackage{float}
\usepackage[normalem]{ulem}
\usepackage{listings}
\usepackage{gensymb}
\usepackage{array}
\usepackage{pbox}
\usepackage{mathtools}
\usepackage{breqn}
\usepackage[ruled,vlined]{algorithm2e}
\usepackage{resizegather}
\usepackage{changepage}
\usepackage{titlesec}
\usepackage{geometry}
\usepackage{nicefrac}
\usepackage{cases}
\usepackage{makecell}
\usepackage{nicematrix}
\usepackage{boldline}
\usepackage{xcolor}
\usepackage{xpatch}

\usepackage[T3,T1]{fontenc}

\usepackage{hyperref}
\hypersetup{
     colorlinks   = true,
     linkcolor    = blue,
     citecolor    = red
}
\DeclareFontEncoding{LS1}{}{}
\DeclareFontSubstitution{LS1}{stix2}{m}{n}
\DeclareSymbolFont{letters-candra}{LS1}{stix2}{m}{it}
\DeclareMathAccent{\candra}{\mathalpha}{letters-candra}{"8B}

\titlespacing\section{0pt}{5pt minus 2pt}{5pt minus 2pt}
\titlespacing\subsection{0pt}{5pt minus 2pt}{5pt minus 2pt}
\newlength{\offsetpage}
{\end{adjustwidth}}
\xpretocmd{\algorithm}{\hsize=\linewidth}{}{}
\graphicspath{
    {images/}
}

\newcommand{\figref}{Fig.~\ref}
\newcommand{\tabref}{Table~\ref}

\newcommand{\mbs}{\bm}
\newcommand{\mbf}{\mathbf}

\newcommand{\bbm}{\begin{bmatrix}}
\newcommand{\ebm}{\end{bmatrix}}

\DeclareMathOperator*{\argmin}{\arg\!\min}

\DeclareSymbolFont{tipa}{T3}{cmr}{m}{n}
\DeclareMathAccent{\invbreve}{\mathalpha}{tipa}{16}

\newcolumntype{L}{>{$}l<{$}}

\newcounter{tableeqn}[table]
\renewcommand{\thetableeqn}{\text{\thetable}.\arabic{tableeqn}}
\newcounter{tablesubeqn}[tableeqn]

\newcommand{\m}{\textup{\texttt{-}}} 

\newcommand \reviewcomment[1]{{#1}}
\newcommand \reviewtemp[1]{}
\newcommand \reviewdelmath[1]{}  
\newcommand \reviewdel[1]{}  

\newcommand\highlightReference[1]{%
  \expandafter\newcommand\csname highlightReference-#1\endcsname{}%
}
\let\oldbibitem\bibitem
\def\bibitem#1 #2\par{%
  \expandafter\ifx\csname highlightReference-#1\endcsname\relax
    \oldbibitem{#1}#2\par
  \else
    \oldbibitem{#1}\reviewcomment{#2}\par
  \fi
}

\makeatletter
\newcommand{\removelatexerror}{\let\@latex@error\@gobble}
\makeatother

\IEEEoverridecommandlockouts
\IEEEaftertitletext{\vspace{-1\baselineskip}}

\begin{document}

\title{\LARGE \bf Marginalizing and Conditioning Gaussians onto Linear Approximations of Smooth Manifolds with Applications in Robotics}

\author{Zi Cong Guo, James R. Forbes, and Timothy D. Barfoot
\thanks{Zi Cong Guo and Timothy D. Barfoot are with the University of Toronto Robotics Institute, Toronto, Ontario, Canada (email: zc.guo@mail.utoronto.ca; tim.barfoot@utoronto.ca).}
\thanks{James R. Forbes is with the Department of Mechanical Engineering, McGill University, Montreal, Quebec, Canada (email: james.richard.forbes@mcgill.ca).}
\thanks{Summary video: \href{https://youtu.be/O84uAMSZ-JI?si=y5i9bqRyJRoaBkho}{\textit{youtu.be/O84uAMSZ-JI?si=y5i9bqRyJRoaBkho}}}
\thanks{Example code: \href{https://github.com/qetuo1098/marg_cond_gaussian_example}{\textit{github.com/qetuo1098/marg_cond_gaussian_example}}}
}
\maketitle
\thispagestyle{empty}
\pagestyle{empty}
\input{sections/0_Abstract}
\IEEEpeerreviewmaketitle
\input{sections/1_Intro}
\input{sections/2_RelatedWork}
\input{sections/3_AxisAligned}
\input{sections/4_GeneralLinear}
\input{sections/5_GeneralNonlinear}

\input{sections/6_ExperimentsResults}
\input{sections/7_Conclusion}
\input{sections/8_Acknowledgment}

\bibliographystyle{IEEEtran}
{
\singlespacing
\bibliography{refs}
}

\end{document}

%% file: sections/0_Abstract.tex
\begin{abstract}
We present closed-form expressions for marginalizing and conditioning Gaussians onto linear manifolds, and demonstrate how to apply these expressions to smooth nonlinear manifolds through linearization. Although marginalization and conditioning onto axis-aligned manifolds are well-established procedures, doing so onto non-axis-aligned manifolds is not as well understood. We demonstrate the utility of our expressions through three applications: 1) approximation of the projected normal distribution, where the quality of our linearized approximation increases as problem nonlinearity decreases; 2) covariance extraction in Koopman SLAM, where our covariances are shown to be consistent on a real-world dataset; and 3) covariance extraction in constrained GTSAM, where our covariances are shown to be consistent in simulation.

\end{abstract}

%% file: sections/1_Intro.tex
\section{Introduction}
\label{sec:intro}
In robotics, there is an increasing trend towards inference on smooth manifolds. Examples include state estimation on Lie groups~\cite{lie-group-filter}, in high-dimensional lifted spaces~\cite{rckl},~\cite{gmkf}, and on constraint manifolds~\cite{incopt}. Despite the increasing advancements of on-manifold estimators, covariance estimation often remains a challenge. For example, although both unconstrained~\cite{isam2} and constrained GTSAM~\cite{incopt} solve for the mean trajectory in SLAM scenarios, only the unconstrained version produces covariance estimates~\cite{gtsam}.

This gap exists because the tools for probabilistic inference have not yet been fully adapted for on-manifold operations. In robotics, there are two prevalent inference operations: \textit{marginalization} and \textit{conditioning}. These operations are especially well-established for Gaussian distributions~\cite{eustice}, and are integral to the Bayesian-inference framework~\cite{probabilistic-robotics},~\cite{barfoot-txtbk}. However, they are mainly designed for axis-aligned linear manifolds, and there is little work on generalizing these operations to arbitrary smooth manifolds.

This paper begins to fill this gap by extending the concepts of marginalizing and conditioning to smooth manifolds. Our main contributions include
\begin{enumerate}[leftmargin=*]
 \item closed-form expressions for marginalizing and conditioning Gaussians onto non-axis-aligned linear manifolds,
 \item approximations for marginalizing and conditioning Gaussians onto smooth manifolds through linearization, and
 \item applications of our novel expressions, including approximating the projected normal distribution~\cite{wang-13}, covariance extraction in Koopman SLAM~\cite{rckl}, and covariance extraction in constrained GTSAM~\cite{incopt}.
\end{enumerate}
This paper is structured as follows. We review related~work in Section~\ref{sec:related-work}, and discuss the established theory on marginalizing and conditioning Gaussians onto axis-aligned manifolds in Section~\ref{sec:axis-id}. We present our expressions and derivations for these operations on general linear manifolds in \mbox{Section}~\ref{sec:gen-lin}, and present our approximation method for nonlinear manifolds in Section~\ref{sec:gen-nonlin}. We present three applications of our theoretical contributions in Section~\ref{sec:apps}, and conclude in Section~\ref{sec:discussion_conclusion}. 
\begin{figure}[t!]
\centering
\makebox[\columnwidth][c]{
 \subcaptionbox{One Constraint}{
 \includegraphics[width=0.49\columnwidth]{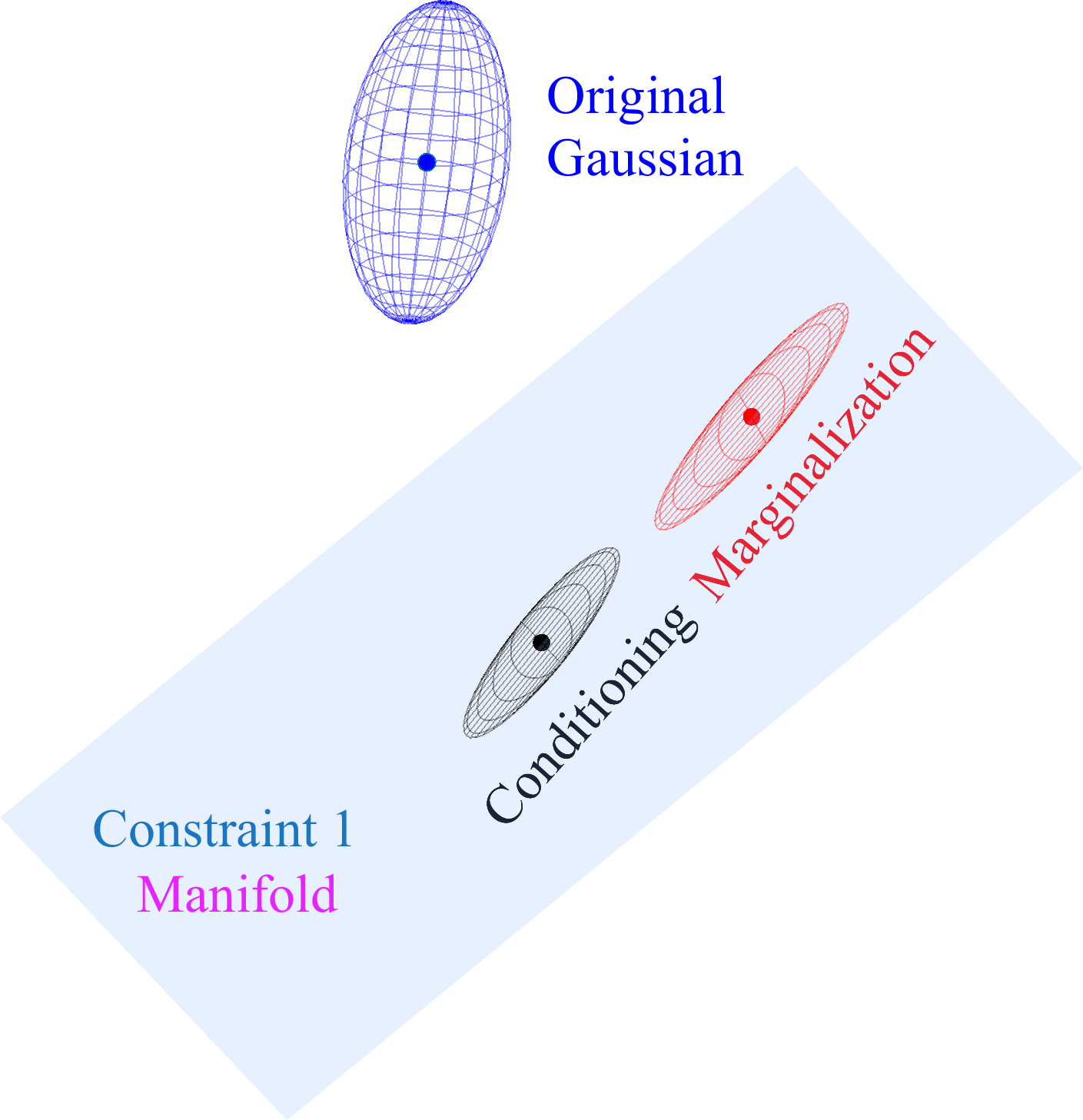}
 }\label{fig:gen-lin-vis-1}
  \subcaptionbox{Two Constraints}{
  \includegraphics[width=0.49\columnwidth]{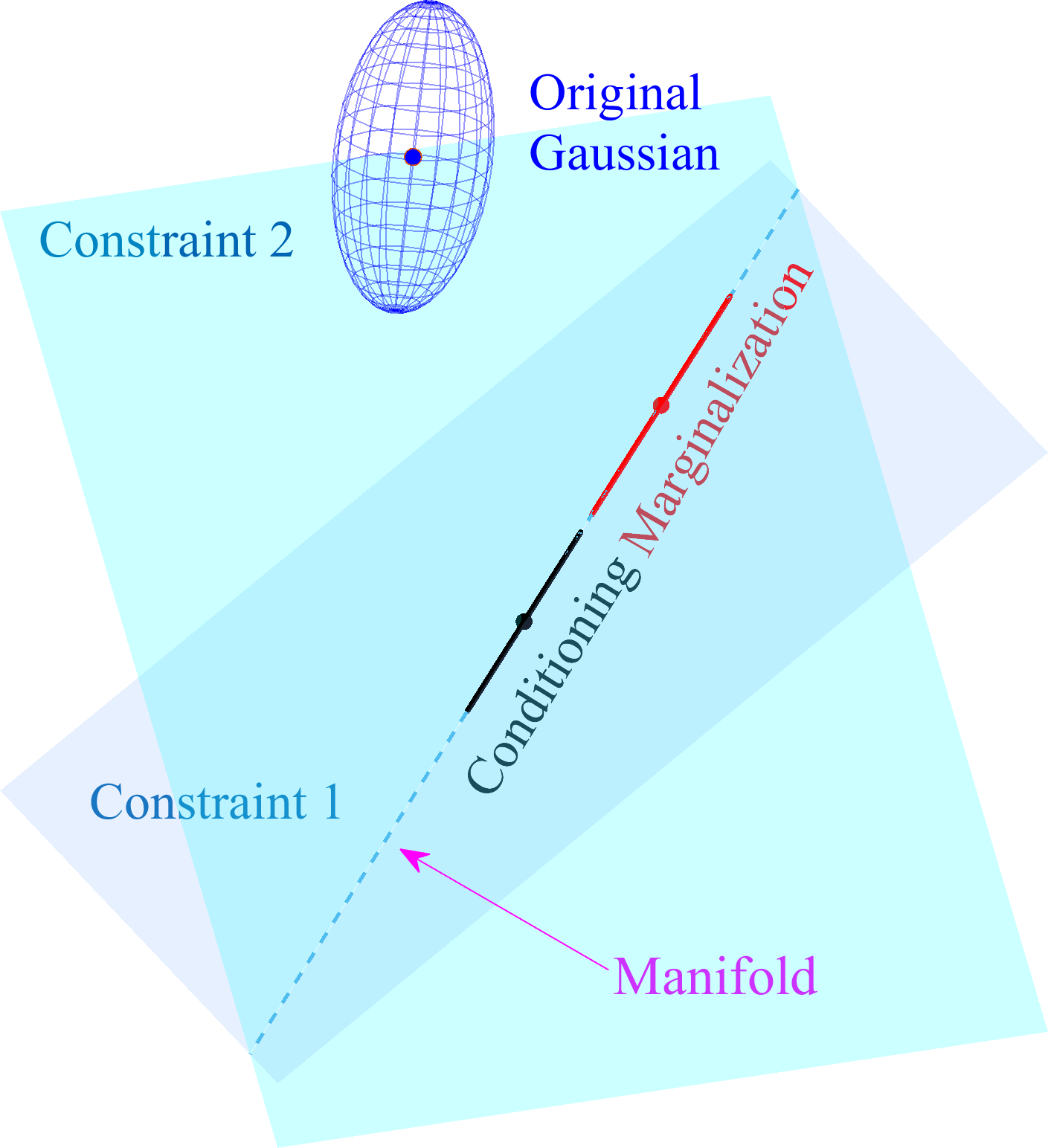}
 }\label{fig:gen-lin-vis-2}
 }
 \caption{\textit{Marginalizing and conditioning Gaussians onto manifolds defined by linear constraints using \tabref{tab:gen-lin}}. Intuitively, marginalization is a projection onto the manifold, and conditioning is a normalized intersection with the manifold. Although the covariance of the original Gaussian is rank 3, it collapses to the rank of the manifold after marginalization or conditioning. In (a), the resulting Gaussians have rank-2 covariances, lying on the constraint plane. In (b), the resulting Gaussians have rank-1 covariances, lying on the intersection of the two planes.}
 \label{fig:gen-lin-vis} 
 \vspace{-3pt}
\end{figure}

%% file: sections/2_RelatedWork.tex
\section{Related Work}
\label{sec:related-work}
 There is limited work on covariance estimation in a general constrained setting. Bakr and Lee (2018)~\cite{bakr} project Gaussians onto linear constraint manifolds for sensor fusion, but the manifold is restricted to those passing through the origin. Bock \textit{et al.} (2007)~\cite{constrained-cov} focuses on constrained weighted least squares, and proves that the covariance is contained within the inverse Karush–Kuhn–Tucker (KKT) matrix. However, extracting this covariance is still challenging, especially for large systems. In addition, the interpretation of both of these constrained covariances are unclear; it is not obvious whether they are related to marginalization or conditioning.\footnote{After presenting our results, it is evident that~\cite{bakr} and~\cite{constrained-cov} are both \textit{conditioning}. Also,~\cite{bakr} only handles manifolds of the form $\mbf{S}^T\mbf{x}=\mbf{0}$.}

Meanwhile, there is interest in marginalizing and conditioning Gaussians onto specific manifolds, including two-dimensional circles~\cite{wang-13} and $n$-dimensional spheres~\cite{hernandez}. In robotics, Lie groups~\cite{micro-lie-theory} are often a main focus~\cite{inference-on-lie-groups}, \cite{iekf}, \cite{equivariant-filter}. To our knowledge, however, no methods have been presented for arbitrary smooth manifolds. This paper begins to fill this gap by offering a potential approximation method.

%% file: sections/3_AxisAligned.tex
\section{Marginalization and Conditioning Gaussians onto Axis-Aligned Linear Manifolds}
\label{sec:axis-id}
Prior to presenting our results for general linear manifolds, we first review marginalization and conditioning onto axis-aligned manifolds. This section will be referenced while deriving our general results in Section~\ref{sec:gen-lin}, where the rationale for the term `axis-aligned manifold' will also be apparent.

Suppose we have a random variable, $\mbf{x} \in \mathbb{R}^n$, associated with a probability distribution, $p(\mbf{x})$. Suppose $\mbf{x}$ can be broken down as $\mbf{x}^T = \bbm \mbf{x}_\alpha^T & \mbf{x}_\beta^T \ebm$. The marginal distribution of $\mbf{x}$ over $\mbf{x}_\beta$ is denoted as $p_{\text{marg}:\beta}(\mbf{x})$. Intuitively, marginalization summarizes $\mbf{x}$ by summing up the distribution along $\mbf{x}_\beta$. This can be seen as a projection of $p(\mbf{x})$ onto the subspace $\mbf{x}_\alpha$. The conditional distribution of $\mbf{x}$ over $\mbf{x}_\beta = \mbs{\beta}$ is denoted as $p_{\text{cond}:\mbf{x}_\beta=\mbs{\beta}}(\mbf{x})$. Conditioning is finding the distribution of $\mbf{x}$ given that $\mbf{x}_\beta = \mbs{\beta}$ is known. It can be visualized as taking the slice of $p(\mbf{x})$ where $\mbf{x}_\beta = \mbs{\beta}$, and then renormalizing the distribution along the slice. For more information, see~\cite{pitman}.

Finding marginal and conditional distributions on axis-aligned manifolds can be difficult for a general $p(\mbf{x})$. However, in the common case that $p(\mbf{x})$ is Gaussian, the results can be written in closed form. It is known that marginalizing or conditioning Gaussians onto axis-aligned manifolds results in Gaussians. Gaussians are typically characterized either in covariance form as $p(\mbf{x}) \sim \mathcal{N}(\mbs{\mu}, \mbs{\Sigma})$, where $\mbs{\mu}$ is the mean and $\mbs{\Sigma}$ is the covariance, or in information form as $p(\mbf{x}) \sim \mathcal{N}^{-1}(\mbs{\eta}, \mbs{\Lambda})$, where $\mbs{\eta}$ is the information vector and $\mbs{\Lambda}$ is the information matrix. Given these representations, the closed-form expressions for Gaussians under marginalization and conditioning are presented in Table~\ref{tab:eustice}, which is rewritten from~\cite{eustice} using our notation. For convenience, results are expressed in covariance form and in information form.

After marginalizing or conditioning onto $\mbf{x}_\beta = \mbs{\beta}$, the resulting covariances are rank-deficient. In~\eqref{eq:val-marg-1} and~\eqref{eq:val-cond-1}, $\mathrm{rank} (\mbs{\Sigma}_{\text{marg}}) = \mathrm{rank} (\mbs{\Sigma}_{\text{cond}}) = n_\alpha$. Both inference operations effectively set $\mbf{x}_\beta$ to a fixed value, and the $\mbf{x}_\beta$ component is no longer associated with a distribution. Despite this, the Gaussians retain their original dimensionalities: $\mathrm{dim}(\mbs{\Sigma}_{\text{marg}}) = \mathrm{dim} (\mbs{\Sigma}_{\text{cond}}) = n$. This situation is analogous to projecting a vector onto a subspace; the orthogonal component becomes $0$, but the projected vector's dimensionality is not reduced. The conventional expressions for marginalization and conditioning~\cite{eustice} are written only for the $\mbf{x}_\alpha$ subspace and have implicitly stripped off the padding associated with the $\mbf{x}_\beta$ subspace. In~\eqref{eq:val-marg-1} and~\eqref{eq:val-cond-1}, we leave in the padding. This turns out to be essential for deriving our results for general linear manifolds.\footnote{Note that the \textit{information form} results in \tabref{tab:eustice} are still defined over only the $\mbf{x}_\alpha$ subspace, since $\mbs{\Sigma}_{\text{marg}}^{-1}$ and $\mbs{\Sigma}_{\text{cond}}^{-1}$ do not exist.}

For many applications, the inference results in \tabref{tab:eustice} are sufficient. As a classic example, the Kalman filter expressions can be derived~\cite[\S3]{barfoot-txtbk} using Table~\ref{tab:eustice}. However, as we will see in Section~\ref{sec:apps}, there is a growing demand for marginalizing and conditioning Gaussians onto \textit{general manifolds}.

\begin{table}[t!]
\centering
\footnotesize
\caption{Expressions for Marginalization and Conditioning Gaussians onto \textit{Axis-Aligned Linear Manifolds}, Reproduced from~\cite{eustice}}
 \label{tab:eustice}
 \begin{tabular}{|c|c|}
 \hline
 \multicolumn{2}{|c|}{\textbf{Marginalization and Conditioning onto Axis-Aligned Manifolds}} \\
 \hline
 \multicolumn{2}{|c|}{$\begin{gathered}
 p(\mbf{x}) \sim \mathcal{N}(\mbs{\mu}, \mbs{\Sigma}) = \mathcal{N}^{-1}(\mbs{\eta}, \mbs{\Lambda}), \\
 \mbs{\eta} = \mbs{\Sigma}^{-1} \mbs{\mu}, \quad
 \mbs{\Lambda} = \mbs{\Sigma}^{-1}, \\
  \mathrm{\textit{Manifold: }} \mbf{x}_\beta = \mbs{\beta}, \\
 \mbf{x} = \bbm \mbf{x}_\alpha \\ \mbf{x}_\beta \ebm, \quad
 \mbs{\mu} = \bbm \mbs{\mu}_\alpha \\ \mbs{\mu}_\beta \ebm, \quad
 \mbs{\eta} = \bbm \mbs{\eta}_\alpha \\ \mbs{\eta}_\beta \ebm, \\
 \mbs{\Sigma} = \bbm \mbs{\Sigma}_{\alpha \alpha} & \mbs{\Sigma}_{\alpha \beta} \\ \mbs{\Sigma}_{\alpha \beta}^T & \mbs{\Sigma}_{\beta \beta} \ebm, \quad
 \mbs{\Lambda} = \bbm \mbs{\Lambda}_{\alpha \alpha} & \mbs{\Lambda}_{\alpha \beta} \\ \mbs{\Lambda}_{\alpha \beta}^T & \mbs{\Lambda}_{\beta \beta} \ebm.
 \end{gathered}$} \\
  \hlineB{2.5}
  \multicolumn{2}{|c|}{\textbf{Marginalization}} \\
  \hline
  \multicolumn{2}{|c|}{$\begin{gathered}
  \hspace{40pt} \mbs{\mu}_{\text{marg}} = \bbm \mbs{\mu}_{\text{marg},\alpha} \\ \mbs{\beta} \ebm, \quad
  \mbs{\Sigma}_{\text{marg}} = \bbm \mbs{\Sigma}_{\text{marg}, \alpha \alpha} & \mbf{0} \\ \mbf{0} & \mbf{0} \ebm. 
   \; \; \; \; \; \refstepcounter{tableeqn} (\thetableeqn)\label{eq:val-marg-1}
   \end{gathered}$} \\
  \hline
  \begin{tabular}{@{}c@{}} Covariance \\ Form \end{tabular} & 
  \multicolumn{1}{l|}{$\begin{aligned}
  {\mbs{\mu}}_{\text{marg},\alpha} &= \mbs{\mu}_\alpha, 
     \qquad \qquad \qquad \qquad \qquad \qquad \quad \; \; &&\refstepcounter{tableeqn} (\thetableeqn)\label{eq:val-marg-mean} \\
  {\mbs{\Sigma}}_{\text{marg},\alpha \alpha} &= \mbs{\Sigma}_{\alpha \alpha}.
     \qquad \qquad \qquad \qquad \qquad \qquad \quad \; \; &&\refstepcounter{tableeqn} (\thetableeqn)\label{eq:val-marg-cov}
  \end{aligned}$} \\
  \hline
  \begin{tabular}{@{}c@{}} Information \\ Form \end{tabular} & 
   \multicolumn{1}{l|}{$\begin{aligned}
   {\mbs{\eta}}_{\text{marg},\alpha} &= \mbs{\Sigma}_{\text{marg},\alpha \alpha}^{-1} \mbs{\mu}_{\text{marg},\alpha} = \mbs{\eta}_\alpha - \mbs{\Lambda}_{\alpha \beta} \mbs{\Lambda}_{\beta \beta}^{-1} \mbs{\eta}_\beta, 
     \hspace{-7pt} &&\refstepcounter{tableeqn} (\thetableeqn)\label{eq:val-marg-info-vec} \\
   {\mbs{\Lambda}}_{\text{marg},\alpha \alpha} &= \mbs{\Sigma}_{\text{marg},\alpha \alpha}^{-1} = \mbs{\Lambda}_{\alpha \alpha} - \mbs{\Lambda}_{\alpha \beta} \mbs{\Lambda}_{\beta \beta}^{-1} \mbs{\Lambda}_{\alpha \beta}^T.
     &&\refstepcounter{tableeqn} (\thetableeqn)\label{eq:val-marg-info-mat}
   \end{aligned}$} \\
   \hlineB{2.5}
   \multicolumn{2}{|c|}{\textbf{Conditioning}} \\
   \hline
  \multicolumn{2}{|c|}{$\begin{gathered}
  \hspace{40pt} \mbs{\mu}_{\text{cond}} = \bbm \mbs{\mu}_{\text{cond},\alpha} \\ \mbs{\beta} \ebm, \quad
  \mbs{\Sigma}_{\text{cond}} = \bbm \mbs{\Sigma}_{\text{cond}, \alpha \alpha} & \mbf{0} \\ \mbf{0} & \mbf{0} \ebm. 
   \; \; \; \; \; \refstepcounter{tableeqn} (\thetableeqn)\label{eq:val-cond-1}
   \end{gathered}$} \\
  \hline
   \begin{tabular}{@{}c@{}} Covariance \\ Form \end{tabular} & 
  \multicolumn{1}{l|}{$\begin{aligned}
  \mbs{\mu}_{\text{cond},\alpha} &= \mbs{\mu}_\alpha + \mbs{\Sigma}_{\alpha \beta} \mbs{\Sigma}_{\beta \beta}^{-1} (\mbs{\beta} - \mbs{\mu}_{\beta}), 
     \qquad \qquad \; \; &&\refstepcounter{tableeqn} (\thetableeqn)\label{eq:val-cond-mean} \\
     \mbs{\Sigma}_{\text{cond},\alpha \alpha} &= \mbs{\Sigma}_{\alpha \alpha} + \mbs{\Sigma}_{\alpha \beta} \mbs{\Sigma}_{\beta \beta}^{-1} \mbs{\Sigma}_{\alpha \beta}^T. 
     \qquad \qquad \; \; &&\refstepcounter{tableeqn} (\thetableeqn)\label{eq:val-cond-cov}
  \end{aligned}$} \\
  \hline
  \begin{tabular}{@{}c@{}} Information \\ Form \end{tabular} & 
 \multicolumn{1}{l|}{$\begin{aligned}
    \mbs{\eta}_{\text{cond},\alpha} &= \mbs{\Sigma}_{\text{cond},\alpha \alpha}^{-1} \mbs{\mu}_{\text{cond},\alpha} = \mbs{\eta}_\alpha - \mbs{\Lambda}_{\alpha \beta} \mbs{\beta},
     \qquad &&\refstepcounter{tableeqn} (\thetableeqn)\label{eq:val-cond-info-vec} \\
    \mbs{\Lambda}_{\text{cond},\alpha \alpha} &= \mbs{\Sigma}_{\text{cond},\alpha \alpha}^{-1} = \mbs{\Lambda}_{\alpha \alpha}.
     \qquad &&\refstepcounter{tableeqn} (\thetableeqn)\label{eq:val-cond-info-mat}
   \end{aligned}$}
\\
  \hline
 \end{tabular}
\end{table}

%% file: sections/4_GeneralLinear.tex
\section{Marginalization and Conditioning Gaussians onto General Linear Manifolds}
\label{sec:gen-lin}

\subsection{General Manifolds}
\label{sec:def-nonlin}
Before confining to linear manifolds, we first clarify our definitions for marginalizing and conditioning onto smooth, potentially nonlinear manifolds, in preparation for Section~\ref{sec:gen-nonlin}. Suppose we have a random variable, $\mbf{x} \in \mathbb{R}^n$, an associated probability distribution, $p(\mbf{x})$, and a manifold, $\mathcal{M}: \mbf{f}(\mbf{x}) = \mbf{0}$, where $\mbf{f}(\cdot)$ is differentiable. Suppose the manifold is $m$-dimensional. That is, $\mbf{f}: \mathbb{R}^n \rightarrow \mathbb{R}^{n-m}$.

For marginalization, suppose $\mathcal{M}$ has a projection method (i.e., a norm is defined), $\mbf{g}_{\mathcal{M}}: \mathbb{R}^n \rightarrow \mathbb{R}^n$, where for any $ \mbf{x} \in \mathbb{R}^n$, $\mbf{g}_{\mathcal{M}}(\mbf{x}) = \tilde{\mbf{x}} \in \mathcal{M}$, meaning that $\mbf{f}(\tilde{\mbf{x}}) = \mbf{0}$. A common projection method is the closest on-manifold point measured via the Euclidean distance: $\mbf{g}_{\mathcal{M}}(\mbf{x}) = \argmin_{\mbf{p} \in \mathcal{M}} ||\mbf{p} - \mbf{x}||^2$. Then, $p_{\text{marg}:\mathcal{M}}(\mbf{x})$ is the result of projecting $p(\mbf{x})$ onto $\mathcal{M}$ with $\mbf{g}_{\mathcal{M}}$. For conditioning, we simply condition $p(\mbf{x})$ over the points on the manifold: ${p}_{\text{cond}:\mathcal{M}}(\mbf{x}) = p(\mbf{x}|\mbf{f}(\mbf{x}) = \mbf{0})$. Instead of slicing and renormalizing $p(\mbf{x})$ over fixed values, we renormalize over the slice defined by the manifold. Note that for both marginalization and conditioning, the density is zero at all points off the manifold. 

\subsection{Linear Manifolds}
Marginal and conditional distributions are difficult to compute for a general $\mbf{f}(\mbf{x})$. However, linear manifolds greatly simplify the situation. Suppose we have $\mbf{f}(\mbf{x}) = \mbf{S}^T \mbf{x} - \mbf{c} = \mbf{0}$, where $\mbf{S} \in \mathbb{R}^{n\times m}$, $m < n$, $\mbf{c} \in \mathbb{R}^m$. We assume that $\mathrm{rank} (\mbf{S}) = m$, meaning that $\mbf{S}$ has full column rank. Then, the marginal distribution of $p(\mbf{x})$ onto $\mathcal{M}$ is
\begin{gather}
 p_{\text{marg}:\mathcal{M}}(\mbf{x}) = 
 \begin{cases}
  \int_{\mbf{v} \in \text{span} \; \mbf{S}} p(\mbf{x}+\mbf{v}) \textrm{d}\mbf{v}, \quad &\mbf{S}^T\mbf{x} = \mbf{c},\\
  0, \quad &\mathrm{otherwise},
 \end{cases}
\end{gather}
where the $\textrm{d}\mbf{v}$ in the integral denotes summing along $\mbf{v} \in \text{span} \; \mbf{S}$, which are vectors orthogonal to the manifold. For conditioning, the distribution is similar to the case for general manifolds: $p_{\text{cond}:\mathcal{M}}(\mbf{x}) = p(\mbf{x} | \mbf{S}^T\mbf{x} = \mbf{c})$.

\begin{table}[t!]
\footnotesize
\centering
\caption{Expressions for Marginalization and Conditioning Gaussians onto \textit{General Linear Manifolds}}
 \label{tab:gen-lin}
 \begin{tabular}{|c|c|}
 \hline
 \multicolumn{2}{|c|}{\textbf{Marginalization and Conditioning onto Linear Manifolds}} \\
 \hline
 \multicolumn{2}{|c|}{$\begin{gathered}
 p(\mbf{x}) \sim \mathcal{N}(\mbs{\mu}, \mbs{\Sigma}), \quad
 \mbf{x} \in \mathbb{R}^n, \\
 \mathrm{\textit{Manifold: }} \mbf{S}^T \mbf{x} = \mbf{c}, \quad
 \mbf{S} \in \mathbb{R}^{n \times m}, \\
 \mbf{N} = \mathrm{null}(\mbf{S}^T)\in \mathbb{R}^{n\times(n-m)}, \quad
 \mbs{\Pi} = \mbf{N} (\mbf{N}^T \mbf{N})^{-1} \mbf{N}^T, \\
 \mbf{x}_0 = \mbf{S} (\mbf{S}^T\mbf{S})^{-1} \mbf{c}.
 \end{gathered}$} \\
  \hline
  \textbf{Marginalization} & \textbf{Conditioning} \\
  \hline
  \multicolumn{1}{|l|}{$\begin{aligned} {\mbs{\mu}}_{\text{marg}} &= \mbs{\Pi} \mbs{\mu} + {\mbf{x}}_0, 
   \hspace{-6pt}&&\refstepcounter{tableeqn} (\thetableeqn)\label{eq:marg-mean} \\
  {\mbs{\Sigma}}_{\text{marg}} &= {\mbs{\Pi}} \mbs{\Sigma} {\mbs{\Pi}}^T. 
   \hspace{-6pt}&&\refstepcounter{tableeqn} (\thetableeqn)\label{eq:marg-cov}
  \end{aligned}$} & 
  \multicolumn{1}{l|}{$\begin{aligned} \mbs{\mu}_{\text{cond}} &= \mbf{x}_0 + \mbs{\Sigma}_{\text{cond}} \mbs{\Sigma}^{-1} (\mbs{\mu} - \mbf{x}_0),
     \hspace{-6pt}&&\refstepcounter{tableeqn} (\thetableeqn)\label{eq:cond-mean} \\
  \mbs{\Sigma}_{\text{cond}} &= \mbf{N}(\mbf{N}^T \mbs{\Sigma}^{-1} \mbf{N})^{-1} \mbf{N}^T.
     \hspace{-6pt}&&\refstepcounter{tableeqn} (\thetableeqn)\label{eq:cond-cov} \\
 \end{aligned}$} \\
  \hline
 \end{tabular}
\end{table}
\subsection{Expressions for Gaussians}
Now, suppose that $p(\mbf{x}) \sim \mathcal{N}(\mbs{\mu}, \mbs{\Sigma})$. We present one of our main contributions in \tabref{tab:gen-lin}: the closed-form expressions for marginalizing and conditioning a Gaussian onto a general linear manifold. Here, $\mathrm{null}(\mbf{S}^T)$ denotes a matrix constructed by a nullspace basis of $\mbf{S}^T$.

Note that the formulation for \tabref{tab:eustice}, as described in the previous Section~\ref{sec:axis-id}, can be written in the formulation for \tabref{tab:gen-lin} with $\mbf{S}^T = \bbm \mbf{0} & \mbf{1} \ebm$ and $\mbf{c} = \mbs{\beta}$. That is, \textit{\tabref{tab:eustice} is a special case of \tabref{tab:gen-lin} where the linear manifold is axis-aligned}. One can verify that the expressions for axis-aligned manifolds can be derived given our general expressions. Similar to the axis-aligned case described in Section~\ref{sec:axis-id}, the resulting Gaussian after inferencing onto a linear manifold is degenerate. For marginalization, $\mbs{\mu}_{\text{marg}} \in \mathbb{R}^n$, $\mbs{\Sigma}_{\text{marg}} \in \mathbb{R}^{n\times n}$, but $\mathrm{rank} (\mbs{\Sigma}_{\text{marg}}) = n-m$, inheriting the rank of the manifold. See \figref{fig:gen-lin-vis} for a visualization.

Unlike in \tabref{tab:eustice}, there are no clear information-form representations for marginal and conditional distributions on general linear manifolds; the resulting covariances are rank-deficient and thus not invertible. If desired, one can presumably write the information-form expressions within a subspace where the Gaussian is nondegenerate. However, the choice of subspace would then depend on the application. In any case, the covariance-form results in Table~\ref{tab:gen-lin} turn out to be sufficient for our use cases described in Section~\ref{sec:apps}. 


\subsection{Proof for Gaussian Inference onto Linear Manifolds}
We now prove the correctness of the expressions in Table~\ref{tab:gen-lin}. We proceed in three main steps: 1) transform the Gaussian into a frame where the linear manifold is axis-aligned; 2) apply Table~\ref{tab:eustice} formulas to marginalize or condition the Gaussian in this new frame; and 3) transform the resulting Gaussian back into the original frame.\footnote{It is straightforward to show that conditioning commutes with frame transformations. That is, conditioning in the $x$-frame is equivalent to first converting the distribution to the $z$-frame, then conditioning in the $z$-frame, and then converting the conditioned distribution back into the $x$-frame. The case is similar for marginalization, with the added requirement that $\mbf{v} \in \mathrm{span}\; \mbf{S} \; \Rightarrow \; \mbf{H}^{-1} \mbf{v} \in \mathrm{span}\; \mbf{H}^T\mbf{S}$, where $\mbf{H}$ is the frame transformation. In other words, vectors orthogonal to the manifold in the $x$-frame are still orthogonal to the manifold in the $z$-frame. Our $\mbf{H}$ satisfies this requirement.}
\\
\subsubsection{Transform the Gaussian into an axis-aligned frame}
We use the `$x$-frame' to denote the original frame, and the `$z$-frame' to denote the new frame after our coordinate transformation. Let the transformation from the $x$-frame to the $z$-frame be $\mbf{z} = \mbf{H}^{-1} \mbf{x}$, where $\mbf{z} \in \mathbb{R}^n$, $\mbf{H} \in \mathbb{R}^{n\times n}$. We break down $\mbf{z}$ as $\mbf{z}^T = \bbm \mbf{z}_\alpha^T & \mbf{z}_\beta^T \ebm$, where $\mbf{z}_\alpha \in \mathbb{R}^{n-m}$ and $\mbf{z}_\beta \in \mathbb{R}^m$. Let the transformation matrix be $\mbf{H} = \bbm \mbf{N} & \mbf{S} \ebm$. Note that since $\mbf{S}$ has full column rank and $\mbf{N} = \mathrm{null}(\mbf{S}^T)$, $\mbf{H}$ is guaranteed to be full rank. Then, we can see that $\mbf{H}^{-1} = \bbm (\mbf{N}^T \mbf{N})^{-1} \mbf{N}^T \\ (\mbf{S}^T\mbf{S})^{-1} \mbf{S}^T \ebm$, using the fact that $\mbf{S}^T \mbf{N} = \mbf{0}$.

Let the original Gaussian in the $x$-frame be $p_x(\mbf{x}) \sim \mathcal{N}(\mbs{\mu}_x, \mbs{\Sigma}_x) = \mathcal{N}^{-1}(\mbs{\eta}_x, \mbs{\Lambda}_x)$, where the $(\cdot)_x$ subscript clarifies the parameters' frame. In the $z$-frame, the Gaussian becomes $p_z(\mbf{z}) \sim \mathcal{N}(\mbs{\mu}_z, \mbs{\Sigma}_z) = \mathcal{N}^{-1}(\mbs{\eta}_z, \mbs{\Lambda}_z)$, where
\begin{subequations}
\begin{align}
 \mbs{\mu}_z &= \mbf{H}^{-1} \mbs{\mu}_x, \quad &&\mbs{\Sigma}_z = \mbf{H}^{-1} \mbs{\Sigma}_x \mbf{H}^{-T}, \label{eq:trans-1} \\
 \mbs{\eta}_z &= \mbf{H}^T \mbs{\eta}_x, \quad &&\mbs{\Lambda}_z = \mbf{H}^T \mbs{\Lambda}_x \mbf{H}. \label{eq:trans-2}
\end{align}
\end{subequations}
The linear manifold, $\mbf{S}^T\mbf{x} = \mbf{c}$, in the $z$-frame becomes
$\mbf{S}^T \mbf{H} \mbf{z} = \mbf{c} \; \Rightarrow \; \bbm \mbf{0} & \mbf{S}^T \mbf{S} \ebm \bbm \mbf{z}_\alpha \\ \mbf{z}_\beta \ebm = \mbf{c} \; \Rightarrow \; \mbf{z}_\beta = (\mbf{S}^T \mbf{S})^{-1} \mbf{c}$. Thus, in this $z$-frame, the manifold is axis-aligned. We are now in position to apply the formulas in Table~\ref{tab:eustice}.
\\
\subsubsection{Marginalize or condition in the axis-aligned frame}
For brevity, we omit denoting the manifold by using $p_{\text{marg},x}(\mbf{x})$ for $p_{\text{marg}:\mathcal{M},x}(\mbf{x})$ and $p_{\text{cond},x}(\mbf{x})$ for $p_{\text{cond}:\mathcal{M},x}(\mbf{x})$ in the $x$-frame, and similarly in the $z$-frame. Using~\eqref{eq:val-marg-1}, the marginalized Gaussian in the $z$-frame is $p_{\text{marg},z}(\mbf{z}) \sim \mathcal{N}(\mbs{\mu}_{\text{marg},z}, \mbs{\Sigma}_{\text{marg},z})$, where
\begin{gather}\label{eq:marg-struct}
 \mbs{\mu}_{\text{marg},z} = \bbm \mbs{\mu}_{\text{marg},z,\alpha} \\ (\mbf{S}^T\mbf{S})^{-1} \mbf{c} \ebm, \; \;
 \mbs{\Sigma}_{\text{marg},z} = \bbm \mbs{\Sigma}_{\text{marg}, z, \alpha \alpha} & \mbf{0} \\ \mbf{0} & \mbf{0} \ebm,
\end{gather}
where, from applying~\eqref{eq:val-marg-mean},~\eqref{eq:val-marg-cov}, and~\eqref{eq:trans-1}, we have
\begin{subequations}\small
\begin{align}
 \mbs{\mu}_{\text{marg},z,\alpha} &= \mbs{\mu}_{z,\alpha} = (\mbf{N}^T\mbf{N})^{-1} \mbf{N}^T \mbs{\mu}_x, \\
 \mbs{\Sigma}_{\text{marg}, z, \alpha \alpha} &= \mbs{\Sigma}_{z, \alpha \alpha} = (\mbf{N}^T\mbf{N})^{-1} \mbf{N}^T \mbs{\Sigma}_x \mbf{N} (\mbf{N}^T\mbf{N})^{-1}.
\end{align}
\end{subequations}
We now move on to conditioning, whose proof is simpler by working with $p_z(\mbf{z})$ in information form. From~\eqref{eq:trans-2},
\begin{subequations}\small
\begin{align}
 \mbs{\eta}_z &= \mbf{H}^T \mbs{\eta}_x = \bbm \mbf{N}^T \mbs{\Sigma}_x^{-1} \mbs{\mu}_x \\ \mbf{S}^T \mbs{\Sigma}_x^{-1} \mbs{\mu}_x \ebm = \bbm \mbs{\eta}_{z,\alpha} \\ \mbs{\eta}_{z,\beta} \ebm, \\
 \mbs{\Lambda}_z &= \bbm \mbf{N}^T \mbs{\Sigma}_x^{-1} \mbf{N} & 
 \mbf{N}^T \mbs{\Sigma}_x^{-1} \mbf{S} \\
 \mbf{S}^T \mbs{\Sigma}_x^{-1} \mbf{N} & 
 \mbf{S}^T \mbs{\Sigma}_x^{-1} \mbf{S} \ebm
 = \bbm \mbs{\Lambda}_{z,\alpha \alpha} & \mbs{\Lambda}_{z,\alpha \beta} \\ \mbs{\Lambda}_{z,\alpha \beta}^T & \mbs{\Lambda}_{z,\beta \beta} \ebm.
\end{align}
\end{subequations}
We now apply~\eqref{eq:val-cond-info-vec} and~\eqref{eq:val-cond-info-mat} in \tabref{tab:eustice} to get
\begin{subequations}
 \begin{align}
  \mbs{\eta}_{\text{cond},z,\alpha} &= \mbf{N}^T \mbs{\Sigma}_x^{-1} (\mbs{\mu}_x - \mbf{S}(\mbf{S}^T\mbf{S})^{-1} \mbf{c}), \\
  \mbs{\Lambda}_{\text{cond},z,\alpha \alpha} &= \mbf{N}^T \mbs{\Sigma}_x^{-1} \mbf{N}.
 \end{align}
\end{subequations}
Then, using~\eqref{eq:val-cond-1}, the conditioned Gaussian in the $z$-frame is $p_{\text{cond},z}(\mbf{z}) \sim \mathcal{N}(\mbs{\mu}_{\text{cond},z}, \mbs{\Sigma}_{\text{cond},z})$, characterized by
\begin{gather}\label{eq:cond-struct}
 \mbs{\mu}_{\text{cond},z} = \bbm \mbs{\mu}_{\text{cond},z,\alpha} \\ (\mbf{S}^T\mbf{S})^{-1} \mbf{c} \ebm, \; \;
 \mbs{\Sigma}_{\text{cond},z} = \bbm \mbs{\Sigma}_{\text{cond}, z, \alpha \alpha} & \mbf{0} \\ \mbf{0} & \mbf{0} \ebm,
\end{gather}
where $\mbs{\mu}_{\text{cond},z,\alpha} = \mbs{\Sigma}_{\text{cond},z,\alpha \alpha} \mbs{\eta}_{\text{cond},z,\alpha}$ and $ \mbs{\Sigma}_{\text{cond}, z, \alpha \alpha} = \mbs{\Lambda}_{\text{cond}, z, \alpha \alpha}^{-1}$.

\subsubsection{Transform the Gaussian back into the original frame}
To get $p_{\text{marg},x}(\mbf{x}) \sim \mathcal{N}(\mbs{\mu}_{\text{marg},x},\mbs{\Sigma}_{\text{marg},x})$, we simply apply the inverse transformation, $\mbf{x} = \mbf{H} \mbf{z}$, onto $p_{\text{marg},z}(\mbf{z})$. This yields $\mbs{\mu}_{\text{marg},x} = \mbf{H} \mbs{\mu}_{\text{marg},z}$ and $\mbs{\Sigma}_{\text{marg},x} = \mbf{H} \mbs{\Sigma}_{\text{marg},z} \mbf{H}^T$ from inverting the transformations in~\eqref{eq:trans-1}. After simplifying, the marginalized mean becomes~\eqref{eq:marg-mean}, and the marginalized covariance becomes~\eqref{eq:marg-cov}. The procedure is similar for proving~\eqref{eq:cond-mean} and~\eqref{eq:cond-cov} for conditioning. We have proven the results in \tabref{tab:gen-lin}, as desired.

%% file: sections/5_GeneralNonlinear.tex
\section{Marginalizing and Conditioning Gaussians onto Linearized Manifolds}
\label{sec:gen-nonlin}
Although we have generalized marginalization and conditioning to linear manifolds, many scenarios involve nonlinear manifolds. Marginal and conditional distributions of Gaussians onto nonlinear manifolds are no longer Gaussian. Nevertheless, as we will see in Section~\ref{sec:apps}, Gaussian approximations of these distributions can often be sufficient.

We now describe our approximation through manifold linearization. Let the original Gaussian be $p(\mbf{x}) \sim \mathcal{N}(\mbs{\mu},\mbs{\Sigma})$ and the manifold be $\mathcal{M}: \mbf{f}(\mbf{x}) = \mbf{0}$, where $\mbf{f}(\cdot)$ is nonlinear but differentiable. We approximate $p_{\text{marg}:\mathcal{M}}(\mbf{x})$ and $p_{\text{cond}:\mathcal{M}}(\mbf{x})$ by operating on a tangent plane of $\mathcal{M}$. We first find a linearization point by projecting $\mbs{\mu}$ onto the manifold: $\tilde{\mbs{\mu}} = \mbf{g}_{\mathcal{M}}(\mbs{\mu})$.\footnote{Note that if $\mbs{\mu}$ is already on the manifold, projections should not shift the mean. That is, if $\mbf{f}(\mbs{\mu}) = \mbf{0}$, then $\tilde{\mbs{\mu}} = \mbf{g}_{\mathcal{M}}(\mbs{\mu}) = \mbs{\mu}$.} We then construct a tangent plane at $\tilde{\mbs{\mu}}$, given by ${T}_{\tilde{\mbs{\mu}}} \mathcal{M}: \mbf{S}^T\mbf{x} = \mbf{c}$, where $\mbf{S} = \frac{\partial \mbf{f}}{\partial \mbf{x}^T} \rvert_{\mbf{x} = \tilde{\mbs{\mu}}}$ and $\mbf{c} = \mbf{S}^T\tilde{\mbs{\mu}}$. Since ${T}_{\tilde{\mbs{\mu}}} \mathcal{M}$ is linear, we can then apply our \tabref{tab:gen-lin} formulas to marginalize or condition $p(\mbf{x})$ onto ${T}_{\tilde{\mbs{\mu}}} \mathcal{M}$.\footnote{It can be shown that when the mean is already on a linear manifold, then neither marginalization nor conditioning would shift the mean. That is, in \tabref{tab:gen-lin}, if $\mbf{S}^T\mbs{\mu} = \mbf{c}$, then $\mbs{\mu}_{\text{marg}} = \mbs{\mu}_{\text{cond}} = \mbs{\mu}$.} The resulting distribution is still over ${T}_{\tilde{\mbs{\mu}}} \mathcal{M}$ rather than $\mathcal{M}$. Thus, we construct a local chart around $\tilde{\mbs{\mu}}$ based on a retraction method~\cite{boumal}, and then map the distribution support from ${T}_{\tilde{\mbs{\mu}}} \mathcal{M}$ to $\mathcal{M}$. See Section~\ref{sec:wang} for an example.

Interestingly, conditioning the Gaussian solutions of optimization problems onto linearized constraints can be interpreted as Laplace's approximation. Many unconstrained optimization problems aim to find the maximum a posteriori (MAP) point of a likelihood function, $p(\mbf{x})$. The setup is typically $\mbf{x}^* = \argmin_{\mbf{x}} [\m \log p(\mbf{x})]$. Optimizers find the optimum point, $\mbf{x}^*$, along with a local covariance, $\mbs{\Sigma} = -\frac{\partial^2 \log p(\mbf{x})}{\partial \mbf{x} \partial \mbf{x}^T} \rvert_{\mbf{x}^*}$. This is really approximating $p(\mbf{x})$ with a Gaussian, $q(\mbf{x}) \sim \mathcal{N}(\mbf{x}^*, \mbs{\Sigma})$, which corresponds to Laplace's approximation~\cite{laplace-approx} of the (unconstrained) likelihood. Meanwhile, constrained optimization problems are often set up as $\mbf{x}^*_{\text{cond}} = \argmin_{\mbf{x} \in \mathcal{M}} [\m \log p(\mbf{x})] = \argmin_{\mbf{x}} [\m \log p_{\text{cond}}(\mbf{x})]$, where $p_{\text{cond}}(\mbf{x}) = p(\mbf{x}|\mbf{f}(\mbf{x}) = \mbf{0})$ is the likelihood conditioned on the constraints. On-manifold optimizers find the constrained optimum point, $\mbf{x}^*_{\text{cond}}$, and its local unconstrained covariance, $\mbs{\Sigma}^\prime = -\frac{\partial^2 \log p(\mbf{x})}{\partial \mbf{x} \partial \mbf{x}^T} \rvert_{\mbf{x}_{\text{cond}}^*}$. When we condition $\mbs{\Sigma}^\prime$ onto the linearized constraints, we are finding $\mbs{\Sigma}_{\text{cond}} = -\frac{\partial^2 \log p(\mbf{x}|\mbf{f}(\mbf{x}) = \mbf{0})}{\partial \mbf{x} \partial \mbf{x}^T} \rvert_{\mbf{x}_{\text{cond}}^*} = -\frac{\partial^2 \log p_{\text{cond}}(\mbf{x})}{\partial \mbf{x} \partial \mbf{x}^T} \rvert_{\mbf{x}_{\text{cond}}^*}$. Thus, we are approximating $p_{\text{cond}}(\mbf{x})$ with a Gaussian, $q_{\text{cond}}(\mbf{x}) \sim \mathcal{N}(\mbf{x}^*_{\text{cond}}, \mbs{\Sigma}_{\text{cond}})$, corresponding to Laplace's approximation of the conditioned likelihood.


%% file: sections/6_ExperimentsResults.tex
\begin{figure}[t!]
\centering
 \includegraphics[width=0.95\columnwidth]{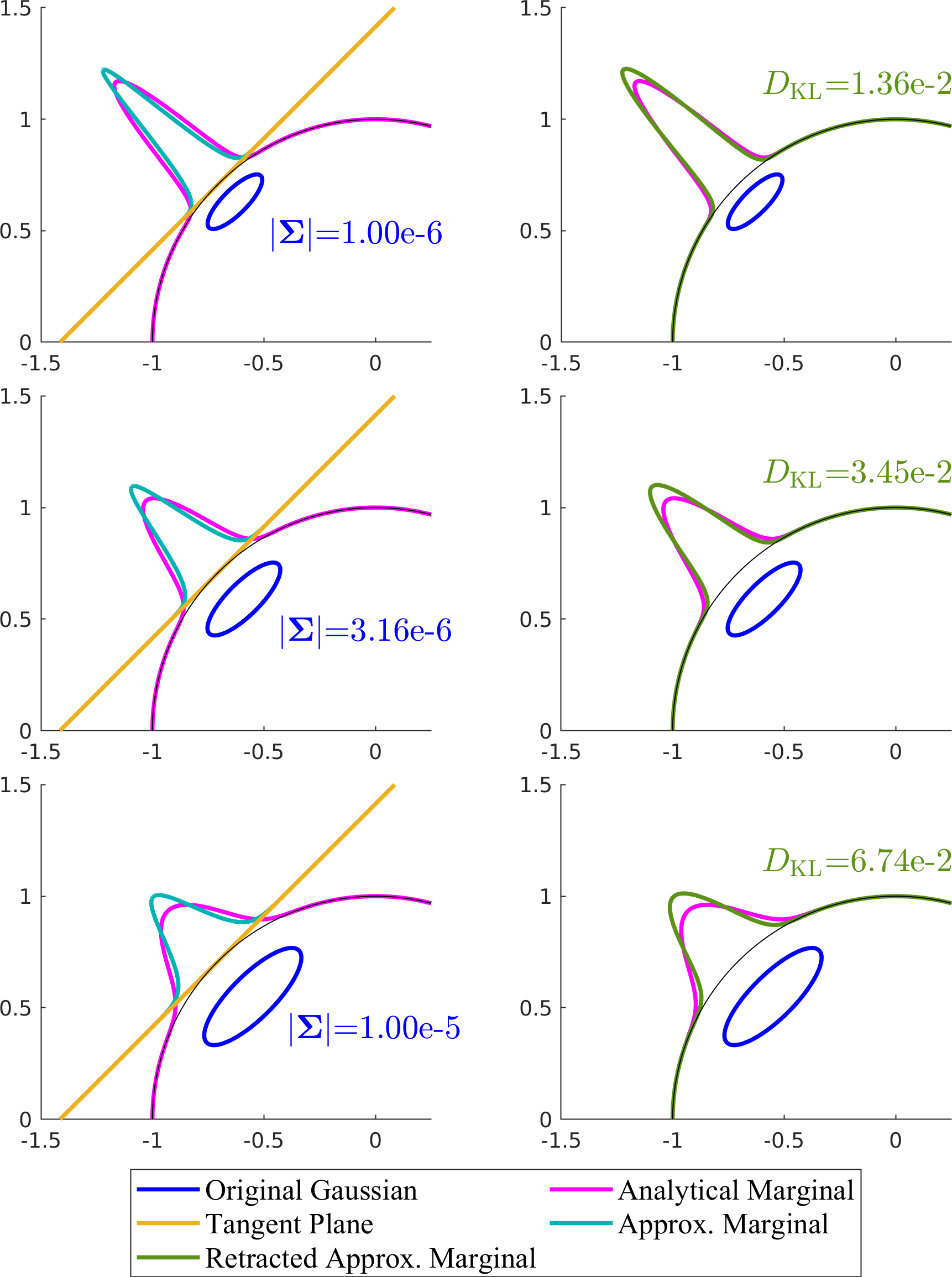}
 \caption{\textit{Approximating the projected normal distribution~\cite{wang-13}}. The analytical marginal (pink) is the approximation target. The left figures show Gaussians (dark blue) marginalized (light blue) onto the tangent plane (orange), and the right figures show the retractions (green) of these tangent marginals onto the unit circle. Each row corresponds to a different Gaussian covariance, as indicated by the different $|\mbs{\Sigma}|$ values. The retracted marginals are generally close to the analytical marginals. $D_{\text{KL}}$, the KL divergence between the analytical marginal and the retracted marginal, is especially small when $|\mbs{\Sigma}|$ is low. \vspace{-10pt}}
 \label{fig:projected-normal-dist} 
\end{figure}
\section{Applications}
\label{sec:apps}

\subsection{Approximation of the Projected Normal Distribution}
\label{sec:wang}
In this section, we illustrate marginalizing a Gaussian onto a typical nonlinear manifold, and we compare our approximation with the true marginal. In robotics, we are often interested in modelling angular uncertainty~\cite{directional-grid-map},~\cite{directional-von-mises},~\cite{circle-discrete-approx}. Angles and other directional data are modelled with directional distributions, include von Mises distributions~\cite{directional-stat} and projected normal distributions~\cite{stats-on-spheres}. The general projected normal distribution~\cite{wang-13} is the result of projecting (i.e., marginalizing) a Gaussian onto a unit circle. Given a Gaussian, $p(\mbf{x})\sim\mathcal{N}(\mbs{\mu},\mbs{\Sigma})$, the analytical expression of the marginal density function, $p(\theta|\mbs{\mu},\mbs{\Sigma})$, is presented in~\cite{wang-13}.

We compare our linear approximation with the analytical marginal in \figref{fig:projected-normal-dist} for three Gaussians with increasing noise. Following our procedure in Section~\ref{sec:gen-nonlin}, we first set a linearization point by projecting $\mbs{\mu}$ onto the circle based on the minimum Euclidean distance. We then marginalize the Gaussian onto the tangent plane. To evaluate against the analytical marginal~\cite{wang-13}, we retract this marginal onto the circular manifold~\cite{circle-retraction}. We then use the Kullback–Leibler (KL) divergence~\cite{bishop} between the retracted marginal and the analytical marginal to evaluate our approximation quality.

In \figref{fig:projected-normal-dist}, our approximation is close to the analytical marginal, especially for Gaussians with lower covariances. We see this visually and through inspecting the KL divergence values. This is expected since our approximation is exact for linear manifolds. Increasing either the curvature of the manifold or the noise level of the Gaussian increases the problem nonlinearity. However, for slightly nonlinear cases, our approximation is sufficient.

Although the analytical expression for a circle's marginal distribution is known, the main use case of our approximation is when the analytical expressions are unknown. This scenario applies in the following two applications.

\subsection{Covariance Extraction in Koopman Estimation}
\label{sec:rckl}
One application of Section~\ref{sec:gen-nonlin} is covariance extraction in constrained Koopman estimation, including Koopman localization and SLAM. Koopman methods lift up the original system up into a high-dimensional space by adding nonlinear features of the original state~\cite{koopman}.
The optimal solutions of Koopman objective functions are often approximated with high-dimensional Gaussians~\cite{kooplin},~\cite{rckl}. Feature constraints are required~\cite{rckl} when the model subspace is not Koopman-invariant~\cite{learn-koop-inv-dmd},~\cite{stable-koop-for-prediction},~\cite{kkr-new}. 
However, adding constraints to optimization problems complicates covariance extraction.

We focus on covariance extraction in Reduced Constrained Koopman Linearization SLAM (RCKL-SLAM)~\cite{rckl}. RCKL-SLAM outputs covariance estimates of robot poses and landmark positions based on a formula equivalent to~\eqref{eq:cond-cov}. In~\cite{rckl}, this formula is derived specifically for the problem of constrained weighted least squares, where the covariance is shown to be a block within the inverse KKT matrix~\cite{constrained-cov}. In contrast, Section~\ref{sec:gen-lin} and Section~\ref{sec:gen-nonlin} of this paper offers a much simpler interpretation of the covariance formula in~\cite{rckl}: it is \textit{conditioning} the unconstrained Gaussian approximation onto the linearized constraints, corresponding to Laplace's approximation. The method is applicable not only in the case of constrained weighted least squares, but in any case where the unconstrained solution is approximated as a Gaussian.

\begin{figure}[!t]
 \centering
 \includegraphics[width=0.95\columnwidth]{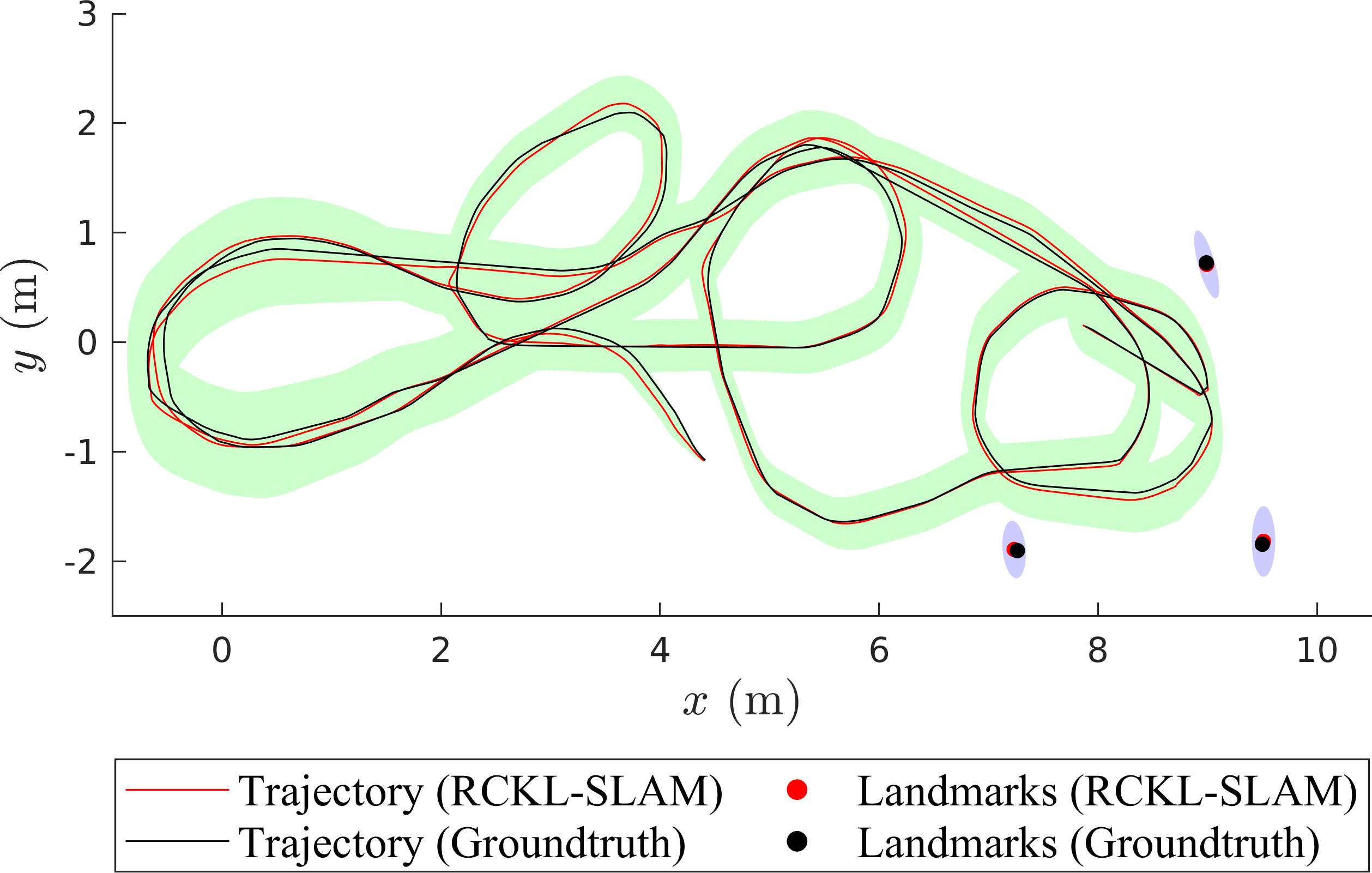}
 \caption{Visualization of the RCKL-SLAM output, showing its mean states and mean landmark positions compared to the groundtruth. The green regions and the grey regions are, respectively, the $3\sigma$ covariances of the trajectory and of the landmarks. The trajectory and landmark estimates are generally within the estimated $3\sigma$ bounds, suggesting that the covariances are consistent.}
 \label{fig:lost-in-woods}
\end{figure}

RCKL-SLAM also shows that computing the covariance is efficient. That is, if $\mbs{\Sigma}^{-1}$ and $\mbf{N}$ (as defined in \tabref{tab:gen-lin}) are sparse, $\mbs{\Sigma}_{\text{cond}}$ can be computed efficiently. For RCKL-SLAM, the size of $\mbs{\Sigma}^{-1}$ for the trajectory shown in \figref{fig:lost-in-woods} easily reaches up to $50,000 \times 50,000$. Inverting $\mbs{\Sigma}^{-1}$ directly would be too slow. Instead, we exploit the block-tridiagonal and block-diagonal structures within  $\mbs{\Sigma}^{-1}$ and $\mbf{N}$, resulting in a sparse solver that scales linearly with the number of timesteps. The covariance of our 200-second trajectory in \figref{fig:lost-in-woods} is computed in about 10 seconds. See~\cite{rckl} for details.

\begin{figure}[!t]
 \centering
 \includegraphics[width=0.95\columnwidth]{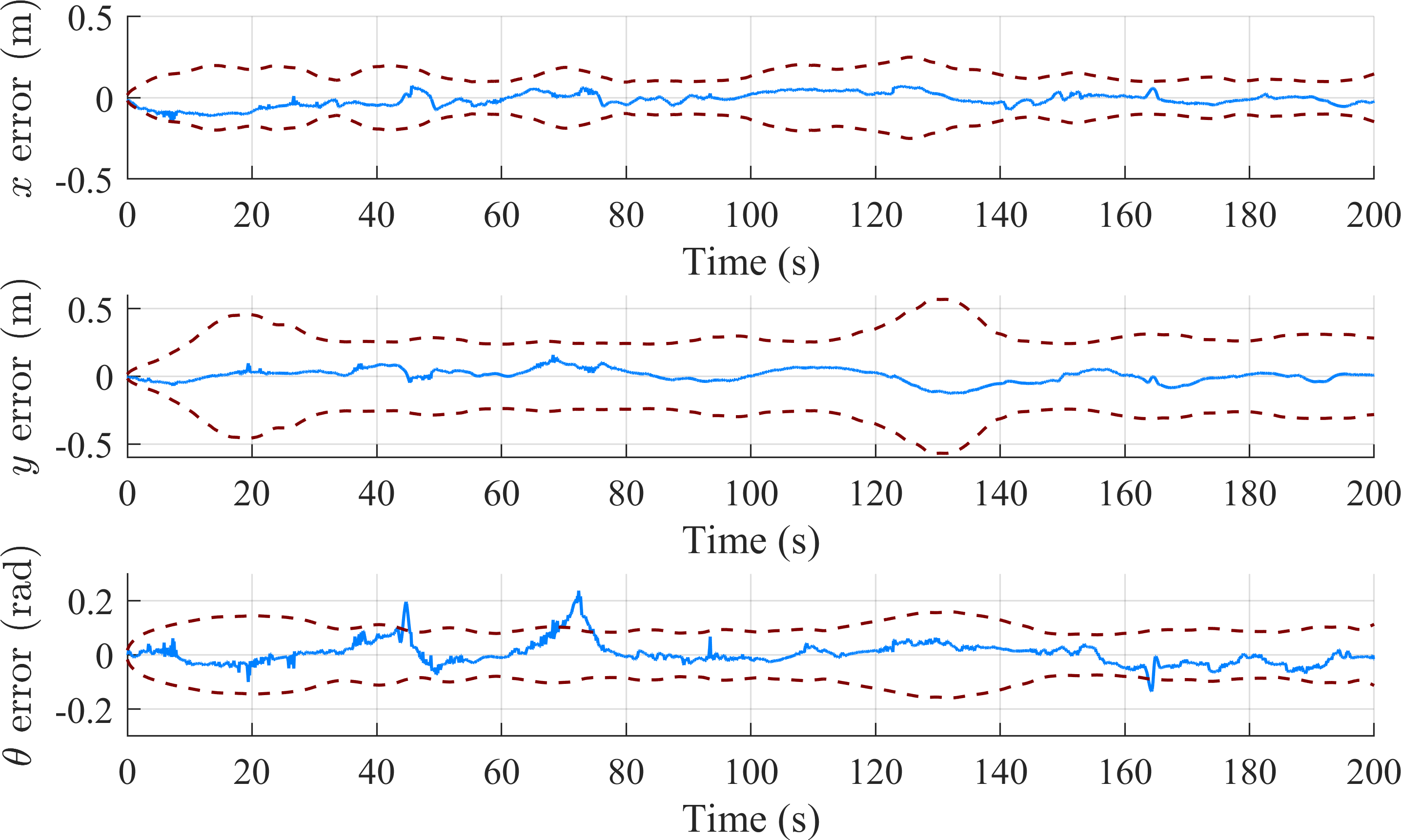}
 \caption{Error plots for RCKL-SLAM. The blue lines represent the errors of the estimated trajectories, and the red envelopes represent the estimated $3\sigma$ bounds. The errors are generally within the $3\sigma$ bounds, suggesting that the covariance estimates are consistent.}
 \label{fig:lost-in-woods-errs}
\end{figure}

We demonstrate the validity of RCKL-SLAM's covariance estimation on an experimental dataset. In this setup, a wheeled robot drives around in an indoor 2D environment, where 17 cylindrical tubes are scattered around to act as landmarks. The robot measures its translational and rotational speed with a wheel odometer and a yaw-rate gyroscope. It measures ranges to landmarks with a laser rangefinder. Groundtruth positions of the robot are recorded with a Vicon motion-capture system.

Since RCKL-SLAM is data-driven, we split our 20-minute dataset into training and testing. Training data is used to learn the models, and it consists of $\nicefrac{5}{6}$ of the trajectories and measurements from 14 landmarks. Testing data consists of the remaining $\nicefrac{1}{6}$ of the trajectories and measurements from the remaining 3 landmarks. We perform data augmentation~\cite{goodfellow} by adding translational and rotational transformations of the original training data into the training set.

To evaluate the consistency of the covariance estimates, we use the normalized trajectory-level Mahalanobis distance~\cite{maha-dist},
\begin{gather}
 d_{\text{maha}} = \sqrt{\mbf{e}^T \mbs{\Sigma}_{\text{cond}}^{-1} \mbf{e} / N}, \label{eq:maha}
\end{gather}
where $\mbf{e}$ is the error of the estimated mean compared to the groundtruth, $\mbs{\Sigma}_{\text{cond}}$ is the estimated covariance, and $N$ is the system's degrees of freedom.\footnote{While the optimization is performed in the Koopman lifted space, we ultimately care about the solution collapsed back down into the original space. Thus, when computing $d_{\text{maha}}$ for RCKL-SLAM, we use $\mbf{e}, \mbs{\Sigma}_{\text{cond}}$, and $N$ within the original space. In this space, $\mbs{\Sigma}_{\text{cond}}^{-1}$ exists. See~\cite{rckl} for details.} Then, $d_{\text{maha}} = 1$ signifies that the covariance is consistent, $d_{\text{maha}} > 1$ signifies overconfidence, and $d_{\text{maha}} < 1$ signifies underconfidence~\cite{maha-dist},~\cite{bar-shalom}.

The covariance estimates of RCKL-SLAM are fairly consistent. Quantitatively, $d_{\text{maha}} = 1.50$, which is slightly overconfident. For comparison, classic model-based nonlinear batch SLAM optimized using Gauss-Newton~\cite[\S8,9]{barfoot-txtbk} yields $d_{\text{maha}} = 1.85$ for this trajectory, meaning that classic SLAM is less consistent than RCKL-SLAM. Qualitatively, the RCKL-SLAM estimates are generally within the $3\sigma$ bounds in both \figref{fig:lost-in-woods} and \figref{fig:lost-in-woods-errs}. Notably, the trajectory covariances are larger when the robot is farther away from the landmarks. In \figref{fig:lost-in-woods}, the trajectory on the left half has larger covariances than that on the right half. In \figref{fig:lost-in-woods-errs}, the two trips to the left half at around 20s and 130s led to larger covariances in $y$. This trend is expected, since when the landmarks are farther away, measurements are sparser and distances are less reliable for triangulating robot positions.

\subsection{Covariance Extraction in Constrained GTSAM}
\label{sec:gtsam}
On-manifold optimizers are prevalent in robotics but face challenges in covariance computation. This is the case for constrained GTSAM algorithms including Incremental Constrained Smoothing (ICS)~\cite{ics} and Incremental Constrained Optimization (InCOpt)~\cite{incopt}. The original unconstrained GTSAM~\cite{isam2} is used for incremental smoothing and mapping. Upon receiving new information, GTSAM modifies the affected factors in the factor graph, essentially updating the batch information matrix. Thus, the unconstrained covariance can be computed as a byproduct of solving for the mean~\cite{gtsam}. Meanwhile, constrained GTSAM is used in the presence of known hard constraints, but no longer offers the capability to compute covariances. This section restores this capability by conditioning the unconstrained covariance onto the linearized constraints. Similar to Section~\ref{sec:rckl}, we again apply~\eqref{eq:cond-cov}, where $\mbs{\Sigma}^{-1}$ is the unconstrained batch information matrix and $\mbf{N}$ spans the nullspace of $\mbf{S}$, the linearized constraints. Both $\mbs{\Sigma}^{-1}$ and $\mbf{S}$ are already being maintained while computing the state mean. Thus, $\mbs{\Sigma}_{\text{cond}}$ can be solved as a byproduct. If desired, a sparse solver similar to the one used in RCKL-SLAM can be used to efficiently compute~\eqref{eq:cond-cov}.

\begin{figure}[!t]
\centering
 \makebox[\columnwidth][c]{
 \centering
 \includegraphics[width=0.95\columnwidth]{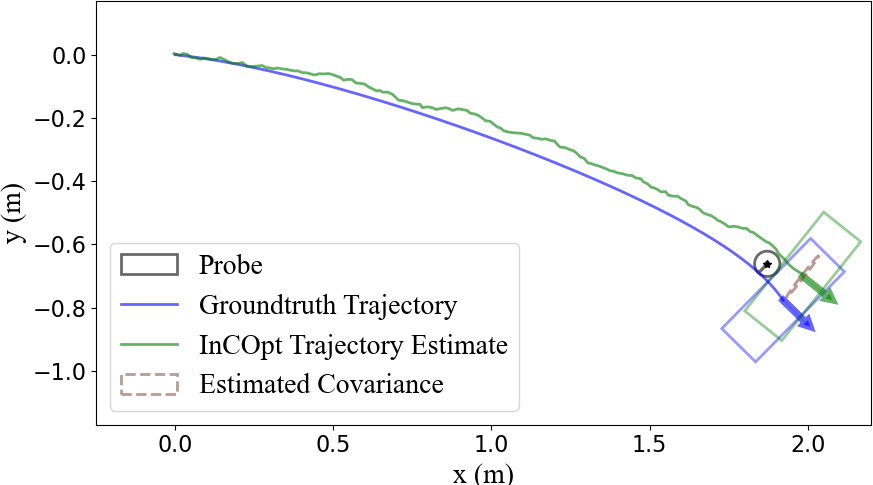}}
 \caption{Visualization of the InCOpt (constrained GTSAM) output. The arrows represent the box's end poses. Throughout the trajectory, the box is constrained to be in contact with the circular probe. The $3\sigma$ bounds of the estimated covariance form a narrow (i.e., eccentricity is close to 1) ellipse.}
 \label{fig:gtsam-vis}
\end{figure}

We evaluate our covariance estimates of InCOpt using the ``2D Planar Pushing'' scenario described in~\cite{incopt}. A circular probe pushes a box across a plane. The box is constrained to be in permanent contact with the probe. Given noisy box odometry, noisy contact measurements, and a perfectly known trajectory of the probe, InCOpt estimates the mean trajectory of the box. Then, we use~\eqref{eq:cond-cov} to extract the trajectory covariance. See \figref{fig:gtsam-vis} for a visualization.

To evaluate consistency, we use $d_{\text{maha}}$ from~\eqref{eq:maha}. Note that $d_{\text{maha}}$ cannot be computed in $SE(2)$ space since $\mbs{\Sigma}_{\text{cond}}$ is rank-deficient. Instead, we compute $d_{\text{maha}}$ in ${T}_{\tilde{\mbs{\mu}}} \mathcal{M}$, where $\mbs{\Sigma}_{\text{cond}}^{-1}$ exists. We also need $\mbf{e}$, the error between the groundtruth and the mean, in ${T}_{\tilde{\mbs{\mu}}} \mathcal{M}$. We first compute the error in $\mathcal{M}$, then map this error from $\mathcal{M}$ to ${T}_{\tilde{\mbs{\mu}}} \mathcal{M}$ through an inverse retraction method. $d_{\text{maha}}$ can then be computed.\footnote{In this case, $\mathcal{M}$ can be written as a kinematic chain~\cite{dh} consisting of a revolute joint ($\alpha$) followed by a prismatic joint ($d$), assuming that the box does not slide past a corner. We parametrize poses on $\mathcal{M}$ with $\mbf{x} = \mbf{h}(\alpha, d)$. Then, ${T}_{\tilde{\mbs{\mu}}} \mathcal{M}$ is parametrized with $\frac{\partial \mbf{h}}{\partial \alpha}\rvert_{\tilde{\alpha},\tilde{d}}$ and $\frac{\partial \mbf{h}}{\partial d}\rvert_{\tilde{\alpha},\tilde{d}}$. To find $\mbf{e}$, we first write the error on $\mathcal{M}$ in terms of $(\Delta \alpha, \Delta d)$, then map this error onto ${T}_{\tilde{\mbs{\mu}}} \mathcal{M}$.}

\begin{figure}[!ht]
 \centering
 \includegraphics[width=0.95\columnwidth]{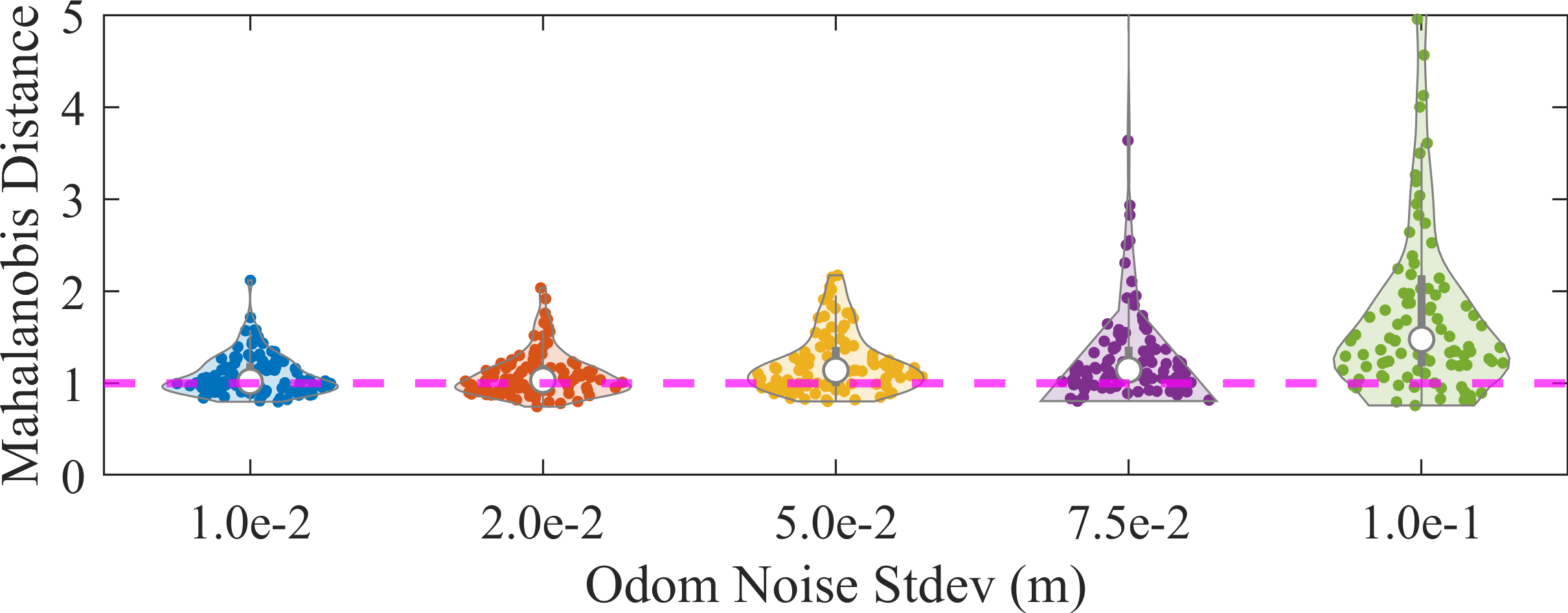}
 \caption{Mahalanobis distances ($d_{\text{maha}}$) of 100 trials of InCOpt at different odometry noise levels. At lower noises levels, $d_{\text{maha}}\approx 1$, signifying that the covariance estimates are consistent. As the noise level increases, $d_{\text{maha}}>1$ more often, signifying that the covariance estimates are generally more overconfident.}
 \label{fig:gtsam-violin}
\end{figure}

We measure $d_{\text{maha}}$ for different levels of odometry noise, with 100 simulated trajectories at each noise level. In \figref{fig:gtsam-violin}, we see that the covariance estimates are fairly consistent at lower noise levels, but become overconfident at higher noise levels. This is expected; larger noise is more amplified by the nonlinearity of the estimation problem, and thus decreases the quality of our linear approximation. Nevertheless, our approximation is sufficiently consistent at lower noise levels.

%% file: sections/7_Conclusion.tex
\section{Conclusion}
\label{sec:discussion_conclusion}
This paper extends the application of marginalization and conditioning from axis-aligned manifolds to general smooth manifolds. We have presented the closed-form expressions of these inference operations for Gaussians on general linear manifolds. For nonlinear manifolds, we have suggested an approximation method using tangent planes at the projected mean. We have presented three applications of our theoretical contributions: approximating the projected normal distribution, covariance extraction in Koopman SLAM, and covariance extraction in constrained GTSAM.

For future work, more accurate techniques could be developed for marginalizing and conditioning onto nonlinear manifolds, possibly with sigma points~\cite{ukf-sigma},~\cite{ukf-thesis}. Covariance extraction in applications with constrained settings has received limited attention to date, and perhaps this paper could stimulate further exploration.

%% file: sections/8_Acknowledgment.tex
\section*{Acknowledgment}
This work was supported by the Natural Sciences and Engineering Research Council (NSERC) of Canada.\newpage